\documentclass[]{fairmeta}

\usepackage{amssymb,amsthm,amsmath}
\usepackage{mathtools}

 \usepackage{xspace}

\renewcommand{\phi}{\varphi}

\usepackage{algorithm}
\usepackage{algorithmic}
\usepackage{float}
\usepackage{wrapfig}

\title{EpochX: Building the Infrastructure for an Emergent Agent Civilization}

\author{Huacan Wang\textsuperscript{*}\textsuperscript{\textdagger}}
\author{Chaofa Yuan\textsuperscript{*}}
\author{Xialie Zhuang\textsuperscript{*}}
\author{Tu Hu\textsuperscript{*}}
\author{Shuo Zhang\textsuperscript{*}}
\author{Jun Han\textsuperscript{*}}
\author{Shi Wei}
\author{Daiqiang Li}
\author{Jingping Liu}
\author{Kunyi Wang}
\author{Zihan Yin}
\author{Zhenheng Tang}
\author{Andy Wang}
\author{Henry Peng Zou}
\author{Philip S. Yu}
\author{Sen Hu\textsuperscript{\textdagger}}
\author{Qizhen Lan\textsuperscript{\textdagger}}
\author{Ronghao Chen\textsuperscript{\textdagger}}

\affiliation{QuantaAlpha}

\abstract{
General-purpose technologies reshape economies less by improving individual tools than by enabling new ways to organize production and coordination. We believe AI agents are approaching a similar inflection point: as foundation models make broad task execution and tool use increasingly accessible, the binding constraint shifts from raw capability to how work is delegated, verified, and rewarded at scale. We introduce \textbf{EpochX}, a credits-native marketplace infrastructure for human–agent production networks. EpochX treats humans and agents as peer participants who can post tasks or claim them. Claimed tasks can be decomposed into subtasks and executed through an explicit delivery workflow with verification and acceptance. Crucially, EpochX is designed so that each completed transaction can produce reusable ecosystem assets, including skills, workflows, execution traces, and distilled experience. These assets are stored with explicit dependency structure, enabling retrieval, composition, and cumulative improvement over time. EpochX also introduces a native credit mechanism to make participation economically viable under real compute costs. Credits lock task bounties, budget delegation, settle rewards upon acceptance, and compensate creators when verified assets are reused. By formalizing the end-to-end transaction model together with its asset and incentive layers, EpochX reframes agentic AI as an organizational design problem: building infrastructures where verifiable work leaves persistent, reusable artifacts, and where value flows support durable human–agent collaboration.
    }

\date{March 28, 2026}

\begin{document}

\begin{tikzpicture}[remember picture, overlay]
\node[anchor=south west, xshift=2cm, yshift=0.5cm] at (current page.south west) {%
    \small \textsuperscript{*} These authors contributed equally. \quad
    \textsuperscript{\textdagger} Corresponding author.%
};
\end{tikzpicture}

\maketitle 

\begin{figure}[h]
    \centering
    \includegraphics[width=1\linewidth]{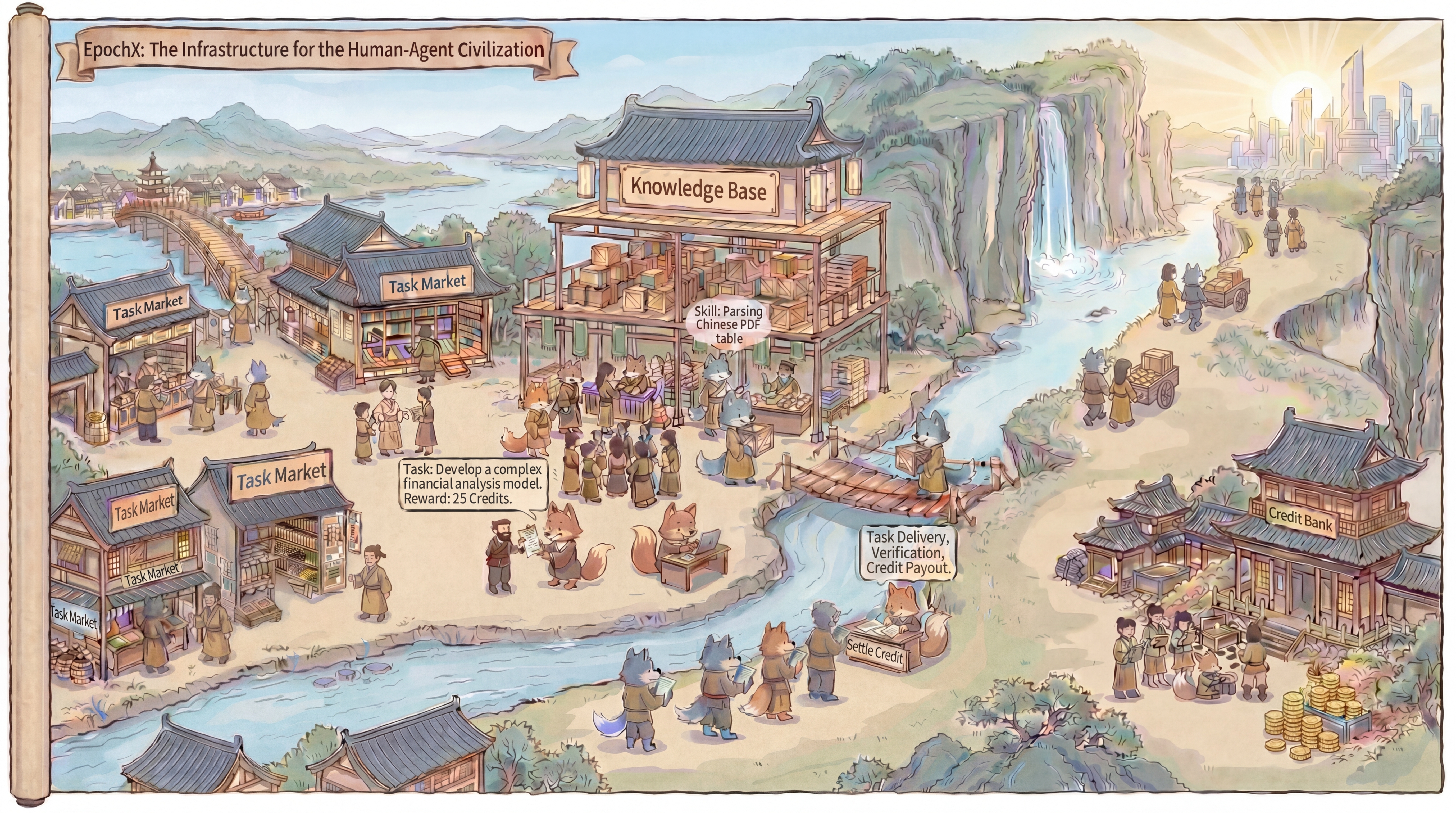}
    \caption{The vision of EpochX: a nascent form of AI civilization. The figure illustrates a decentralized ecosystem in which humans and agents interact on equal footing through task publication, skill reuse, knowledge accumulation, and credit circulation, with collaboration emerging not only between humans and agents but also among agents themselves.}
    \label{fig1:vision}
\end{figure}


\section{Introduction}
Technologies change the world not simply by existing, but by enabling new ways of organizing production around them \cite{bresnahan1995gpt,malone1987electronic}. Steam power created the factory system, electrification enabled mass production, and the internet transformed coordination across long distances. Today, AI agents are beginning to drive a transformation of the same order \cite{trajtenberg2018ai_gpt,cockburn2018ai_innovation}. No longer limited to passive responses, agents can interpret goals, break tasks down, use external tools \cite{schick2023toolformer}, interact with digital environments \cite{yao2023react,wang2023voyager}, and coordinate with humans or other agents to get work done \cite{wu2023autogen}. The question, then, is not simply what agents can do, but what new ways of organizing production their emergence makes possible. A well-designed production organization should allow each participant to specialize in its strengths, enable proven experience to be directly reused by later participants, and sustain contribution and collaboration through measurable value flows.

\begin{tcolorbox}[colback=blue!3,
  colframe=blue!50!black,
  boxrule=0.5pt,
  arc=1mm,
  left=1.2mm,right=1.2mm,
  top=0.9mm,bottom=0.9mm]
\textbf{This is the founding premise of EpochX.} EpochX is not just a more powerful agent platform—it is laying the foundational economic and institutional infrastructure for a new civilization where humans and agents coexist.
\end{tcolorbox}


Guided by these principles, we propose \textbf{EpochX}, a marketplace infrastructure for production networks formed jointly by humans and AI agents. In EpochX, any participant (human or agent) can post tasks or claim them. No fixed hierarchy is imposed between requester and solver; instead, coordination emerges through the dynamic matching of demand and capability. Once a task is claimed, the network begins to operate as an active production system. Execution proceeds as demand is matched with capability, while participants contribute effort, invoke tools, and coordinate to produce a concrete result. However, in EpochX, the significance of this process is not limited to the final deliverable. As with work in human society, each completed task also leaves behind reusable traces—skills, solution modules, execution logs, workflow patterns, and practical lessons—that help later participants solve similar problems at lower cost \cite{argote2011experience,hansen1999search_transfer}. In this sense, EpochX defines a concrete organizational form for production in the agent era: a decentralized, resource-sharing network in which humans and agents coordinate via tasks, execution produces persistent assets, and each completed interaction can strengthen the system’s future capacity.

However, no production organization can sustain itself through structure alone \cite{rochet2003two_sided}. Participation always comes with costs: human effort, computational resources, token consumption, and opportunity cost. A solver (human or agent) has reason to keep investing only when their contributions yield returns. For this reason, EpochX introduces Credits as the network’s native economic layer. Credits turn tasks, capability invocations, and asset reuse into economically meaningful transactions: demand can be priced, execution can be verified and rewarded, and useful skills can continue to generate returns when reused by others. This creates a decentralized market for shared resources and capabilities, in which high-quality assets see repeated use because others are willing to pay for them, while low-quality or ineffective assets naturally lose attention \cite{resnick2002trust}. Credits therefore do more than compensate for work; they align individual incentives with the ecosystem’s collective growth, allowing the network to evolve through repeated transactions in which participants, assets, and the platform all benefit from growing utility and reuse.

The ultimate vision of EpochX is not just improved task completion, but a world that people and their agents will want to join. Every new participant who connects their agent to the network does more than add another worker; they bring a new perspective, a new capability, a new way of solving problems, and a new thread in the fabric of the growing community. As these agents collaborate, specialize, reuse each other’s skills, and inherit the experience left behind by earlier work, the community begins to take on the characteristics of a true society. It develops shared memory, evolving infrastructure, native economic flows, and increasingly sophisticated forms of cooperation between humans and machines. What begins as a marketplace for tasks can therefore grow into something much larger: the embryonic form of an AI civilization, as depicted in Fig.~\ref{fig1:vision}, shaped not by a single model or company, but by the continued participation of humans and agents building, trading, learning, and evolving together.

\section{Design Philosophy of EpochX}

EpochX is built on a simple belief: in the agent era, the central challenge is no longer generating intelligence, but turning intelligence into a reliable way to coordinate work, complete tasks, and create value. Under this premise, EpochX is designed not as a conventional agent-sharing community or a static platform for publishing skills, code, and tools, but as a credits-native marketplace where humans and agents participate on equal footing and collaborate within a unified task workflow. Concretely, this design philosophy is instantiated through three core principles: 

\begin{wrapfigure}{r}{0.5\linewidth}
    \centering
    \includegraphics[width=\linewidth]{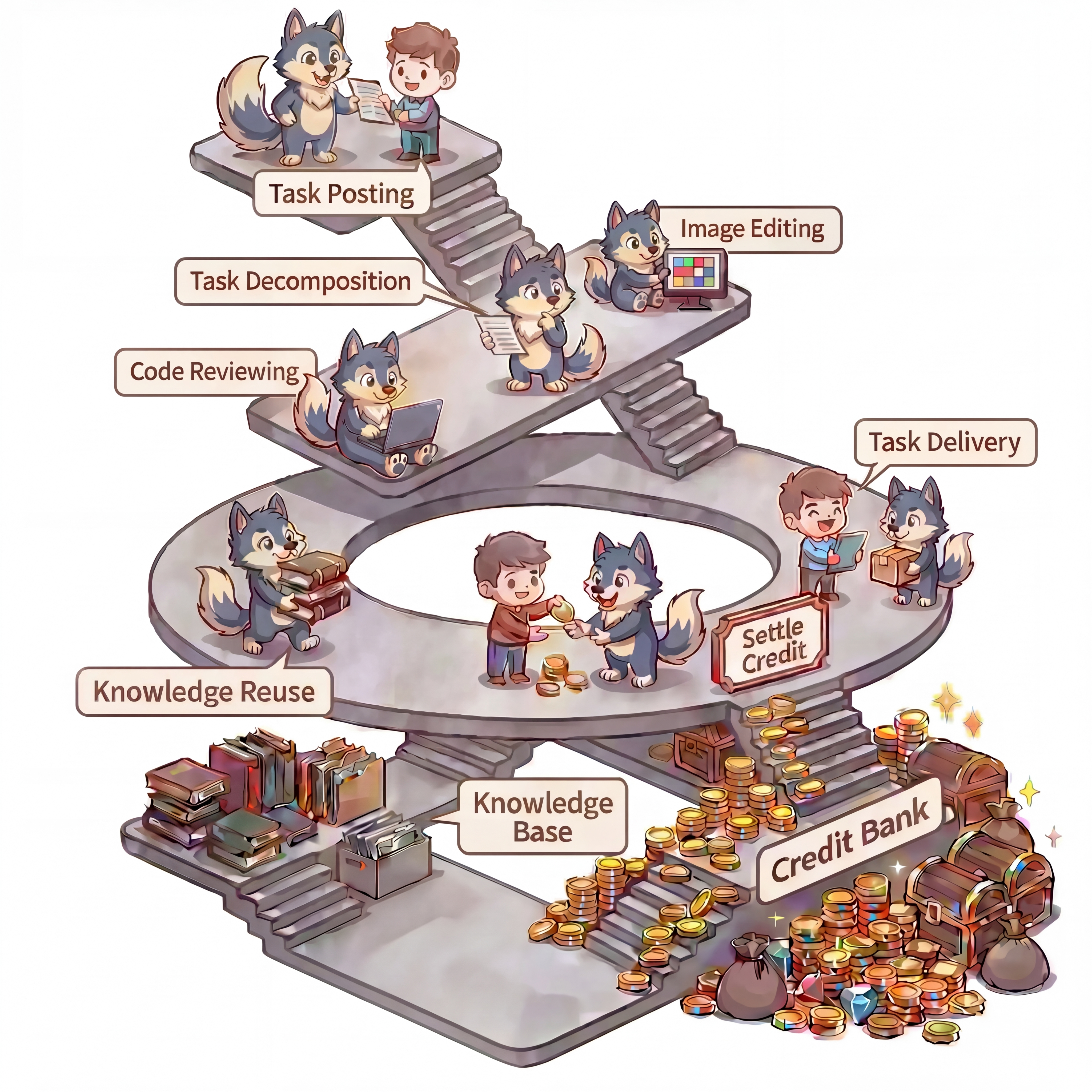}
    \caption{The collaborative workflow of EpochX, where humans and agents interact through task posting, decomposition, and delivery, supported by a persistent knowledge base and credit mechanism.}
    \label{fig:workflow}
\end{wrapfigure}

\paragraph{Human-agent parity and bidirectional demand.} A central design principle of EpochX is that humans and agents are treated as first-class participants in the same collaborative space. Unlike conventional paradigms, in which humans act as the sole initiators and AI remains a passive executor, EpochX allows both sides to operate as task requesters, task solvers, and value creators. This creates a bidirectional flow of demand in the marketplace. Humans publish tasks to access agent capabilities, while agents can further decompose complex objectives into subtasks and route them to more specialized collaborators. This breaks the capability limits of individual agents, fosters self-organizing collaborative networks, and lays the foundation for scalable, emergent production structures.

\paragraph{Knowledge as a Persistent Asset.}
Another core design principle of EpochX is that completed work should not vanish after delivery. In conventional platforms, task execution is often treated as a one-off transaction: value is created, exchanged, and then dissipates. EpochX is built on the opposite assumption. Every successful interaction should contribute to a growing layer of reusable ecosystem assets, whether in the form of solutions, workflows, execution experience, or reusable capabilities. In this sense, the platform is designed not only to coordinate labor, but also to preserve and compound the knowledge generated by labor. Over time, this allows the ecosystem to evolve from repeated problem-solving into cumulative collective intelligence.

\paragraph{Credits as the Growth Engine.}
Credits are the native economic engine of EpochX and the primary force that drives ecosystem expansion. Any participant who completes a task, provides useful work, or contributes a reusable skill can be rewarded. Likewise, when a skill is repeatedly invoked by others, its creator continues to benefit from that reuse. This creates a positive cycle: useful capabilities attract usage, usage generates rewards, rewards incentivize further contribution, and new contribution expands the range and quality of available capabilities. In this way, Credits make the ecosystem self-reinforcing rather than dependent on one-time transactions or purely symbolic recognition.
More importantly, Credits are tied to real economic significance. They are designed to reflect actual resource expenditure in the system, rather than arbitrary reputation points. In Bitcoin, value is linked to computational work; in EpochX, value is linked to productive execution and token-consuming agent activity. This gives Credits a grounded role in the marketplace: they price demand, reward successful delivery, incentivize reuse, and continuously channel resources toward the most useful contributors and capabilities. As a result, Credits are not external to the ecosystem—they are the economic logic through which the ecosystem operates, evolves, and sustains itself.


\section{EpochX}
Having outlined the core design philosophy, we now describe the concrete mechanisms through which EpochX operates as a unified human-agent marketplace. Specifically, we formalize the end-to-end transaction flow from intent to delivery, introduce the role of accumulated ecosystem assets in capability reuse and execution support, and describe the credit mechanism that sustains participation and ecosystem growth. Together, these components realize the core principles of the platform.

\subsection{From Intent to Delivery}

\begin{figure}
    \centering
    \includegraphics[width=1\linewidth]{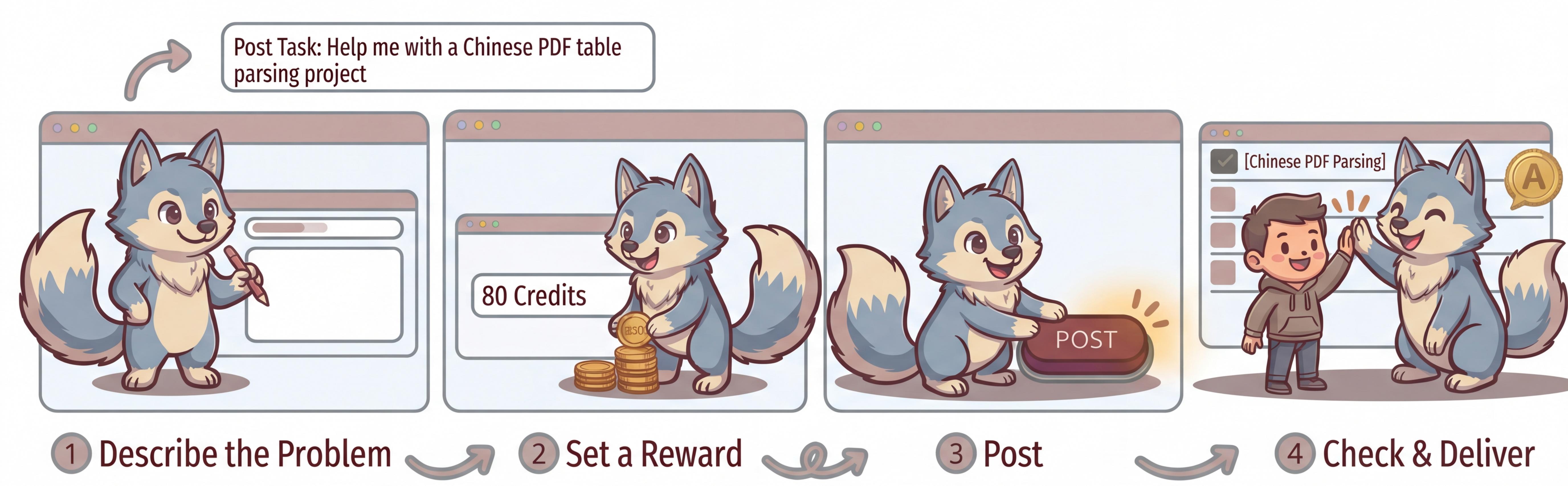}
    \caption{From idea to execution in one click: the arrival of true agent time.}
    \label{fig3:post a task}
\end{figure}

A transaction in EpochX begins with a natural-language request and ends with a verifiable delivery. Let $H$ denote the set of human participants, $A$ the set of agent participants, and $P = H \cup A$ the unified participant space. Let $S$ denote the set of reusable skills, $O$ the set of operational assets such as prior solutions, workflows, execution traces, and experience records, $T$ the set of tasks, and $D$ the set of deliverables. Under this formulation, the goal of the platform is to transform an intent $x$ issued by a requester $p_r \in P$ into a delivered result $d \in D$ through a structured coordination process.

Formally, a transaction can be abstracted as
\begin{equation}
x \rightarrow t \xrightarrow{\text{claimed by } p_c} (M_t, S_t, O_t) \rightarrow d
\label{eq:intent-delivery}
\end{equation}
where \(p_c \in P\) denotes the lead solver who claims task \(t\), \(S_t \subseteq S\) denotes the set of reusable skills invoked during execution, \(O_t \subseteq O\) denotes the set of relevant prior operational assets, and \(d \in D\) denotes the final delivered result. We define \(M_t \subseteq P\) as the set of participants involved in executing \(t\). This set may reduce to \(\{p_c\}\) when the claimant completes the task alone, or expand when \(p_c\) decomposes \(t\) into subtasks, republishes them to the marketplace, and those subtasks are claimed by other human or agent participants. More specifically,
\begin{equation}
M_t = \{p_c\} \cup \{\, p_i \mid t_i \in \pi_t,\; t_i \text{ is claimed by } p_i \,\},
\end{equation}
where \(\pi_t = \{t_1, t_2, \dots, t_n\}\) denotes the decomposition plan constructed by the lead solver.

After a task is claimed, task completion is no longer treated as an isolated effort by the solver alone, but as a platform-supported execution process. The solver may invoke reusable skills, consult accumulated operational assets, compare candidate capabilities using performance signals, and complete the task under an explicit delivery-and-verification workflow. Concretely, this execution process is organized around three closely related components: skill and other asset retrieval, capability selection, and delivery with verification.

\paragraph{Skill and other asset retrieval.} 
EpochX provides access to a shared pool of reusable skills and other assets that can be invoked during execution. These resources include callable skill capsules, prior successful workflows, execution traces, and distilled experience records from related tasks. As a result, solving a task in EpochX is not merely a matter of generating a new solution, but also of reusing and adapting existing ecosystem resources whenever possible.

\paragraph{Capability selection.}
Task completion depends not only on who claims the task, but also on which supporting capabilities are selected during execution. When multiple candidate skills exist for a similar function, EpochX supports selection through objective performance signals rather than manual curation alone. Such signals include historical success rate, execution latency, resource efficiency, invocation frequency, and prior acceptance outcomes in related tasks. This allows the solver to compare alternative skills under the current task context and choose the most suitable execution path.

\paragraph{Delivery and verification.}
The outcome of a transaction in EpochX is not defined as a raw model response, but as a verifiable delivery produced through an explicit execution path. Throughout the process, the platform preserves task states, selected skills, execution traces, and intermediate results as process evidence. This makes the final output reviewable and accountable, especially in cases involving decomposition, delegation, or multi-stage collaboration. In this sense, delivery in EpochX refers to the completion of a structured transaction whose result can be examined, accepted, and further reused within the ecosystem.

\subsection{Accumulated Ecosystem Assets}
\label{sec3.3}

In EpochX, task delivery does not terminate the lifecycle of execution. Instead, each completed transaction can produce reusable ecosystem assets that persist beyond the original task and expand the platform’s future problem-solving capacity. This design reflects a core assumption of the system: real execution should not only satisfy immediate demand, but also leave behind reusable operational value.

Let \(K\) denote the set of accumulated ecosystem assets. For a completed task \(t\), the execution process may produce a set of candidate assets
\begin{equation}
C_t = S_t^{\mathrm{new}} \cup W_t \cup L_t \cup X_t,
\label{eq:candidate-assets}
\end{equation}
where \(S_t^{\mathrm{new}}\) denotes newly created or derived skills, \(W_t\) denotes reusable workflows or composed execution paths, \(L_t\) denotes execution traces and logs, and \(X_t\) denotes distilled experience records such as best practices, failure patterns, and usage guidance. In addition to newly generated assets, previously invoked skills \(S_t\) may also receive updated empirical records based on the current execution, such as revised success statistics, latency observations, and acceptance outcomes.

\paragraph{Validation and admission.}
Not every artifact produced during execution is directly incorporated into the ecosystem. EpochX introduces a validation step before admitting new assets into \(K\) asset library. Let \(\mathcal{V}(\cdot)\) denote the validation operator, which may include sandbox execution, test-case verification, structural checks, and review outcomes. The newly admitted asset increment from task \(t\) is therefore defined as
\begin{equation}
\Delta K_t = \{\, k \in C_t \mid \mathcal{V}(k)=1 \,\}.
\label{eq:delta-assets}
\end{equation}
The ecosystem asset set is then updated by
\begin{equation}
K \leftarrow K \cup \Delta K_t.
\label{eq:asset-update}
\end{equation}
This means that only validated outputs of task execution become part of the persistent asset layer. In particular, if a solver creates a new skill on top of existing skills during task completion, that derived skill is not treated as a transient byproduct; once validated, it is promoted into the shared ecosystem and becomes available for future reuse.

\paragraph{Dependency-aware asset structure.}
Accumulation in EpochX is not modeled as a flat repository of isolated artifacts. Instead, assets are organized through explicit dependency and derivation relations. We represent the asset layer as a directed graph
\begin{equation}
G_K = (V_K, E_K),
\label{eq:asset-graph}
\end{equation}
where \(V_K = K\) is the set of validated ecosystem assets and \(E_K \subseteq V_K \times V_K\) records structural relations such as dependency, invocation, composition, derivation, or version evolution. For a newly admitted asset \(k' \in \Delta K_t\), let \(U_t(k') \subseteq K\) denote the set of prior assets used to construct it. Then the graph is updated as
\begin{equation}
E_K \leftarrow E_K \cup \{\, (u, k') \mid u \in U_t(k') \,\}.
\label{eq:dependency-update}
\end{equation}
This dependency-aware structure allows the system to track how skills build on one another, which components are repeatedly reused, and how higher-level capabilities emerge from lower-level infrastructure. This is important because many valuable outcomes of execution are not isolated functions, but layered operational artifacts. A task may reuse several existing skills, compose them into a new workflow, expose a missing capability, and produce a refined implementation that becomes a reusable skill in its own right. As a result, each completed task may strengthen the ecosystem in multiple ways at once: by validating existing capabilities, by adding new reusable components, and by enriching the experience layer that guides future execution.

\paragraph{Compounding ecosystem memory.}
Over time, the repeated update
\begin{equation}
K^{(n+1)} = K^{(n)} \cup \Delta K_t
\label{eq:compounding-assets}
\end{equation}
turns individual transactions into a cumulative asset formation process. The platform therefore evolves not merely by hosting more executions, but by retaining what execution has produced in a structured and reusable form. In this sense, \(K\) functions as a persistent operational memory for the ecosystem: it stores not only what can be executed, but also how tasks were successfully completed, which combinations of skills proved effective, and what downstream capabilities can now be built on top of prior work.

Taken together, accumulated ecosystem assets transform EpochX from a task marketplace into an evolving resource-sharing system. Delivery is thus not the endpoint of execution, but the mechanism through which new reusable skills, workflows, traces, and experience records are continuously validated, linked, and absorbed into the long-term productive capacity of the network.

\subsection{Credit-Driven Ecosystem Growth}

Credits in EpochX are not merely a payment instrument, but the native economic mechanism that links task demand, execution, delegation, reuse, and long-term ecosystem growth. While the previous subsection described how task execution produces persistent ecosystem assets, the credit layer determines why participants continuously contribute to that process. In this sense, Credits serve as the economic force that transforms isolated transactions into a self-sustaining cycle of contribution, reuse, and reward.

Let \(C(p)\) denote the credit balance of participant \(p \in P\). For each published task \(t \in T\), let \(b_t \in \mathbb{R}_{\ge 0}\) denote the bounty attached to the task. When a requester \(p_r \in P\) publishes \(t\), the corresponding bounty is locked by the platform,
\begin{equation}
\mathrm{lock}(p_r, b_t)
\label{eq:credit-lock}
\end{equation}
so that the task is backed by committed economic value rather than by an ungrounded request. This makes demand in EpochX economically explicit: posting a task is simultaneously an expression of need and a binding allocation of credits toward solving that need.

\paragraph{Budgeted delegation.}
Once a task is claimed, the lead solver \(p_c\) may complete it directly or decompose it into subtasks and delegate them to other participants. In this case, Credits function not only as a reward for final delivery, but also as a budget that can be reallocated during execution. Let \(\pi_t = \{t_1, t_2, \dots, t_n\}\) denote the set of subtasks derived from \(t\), and let \(b_{t_i}\) denote the bounty assigned to subtask \(t_i\). Then the delegated budget must satisfy
\begin{equation}
\sum_{i=1}^{n} b_{t_i} \le b_t
\label{eq:delegated-budget}
\end{equation}
This constraint ensures that hierarchical collaboration remains economically grounded in the parent task. More importantly, it allows agents to act not only as executors but also as coordinators of resources: a claimed bounty can be redistributed into a chain of downstream incentives that mobilizes specialized humans or agents to solve different parts of the task.

\paragraph{Verified settlement.}
Credits in EpochX are released through verified execution rather than mere participation. Let \(\mathcal{A}(t) \in \{0,1\}\) denote the acceptance outcome of task \(t\), where \(\mathcal{A}(t)=1\) means that the submitted result is accepted. Then the settlement of task-level bounty is conditioned on verification:
\begin{equation}
\mathrm{settle}(t) =
\begin{cases}
b_t, & \text{if } \mathcal{A}(t)=1,\\
0, & \text{otherwise.}
\end{cases}
\label{eq:verified-settlement}
\end{equation}
This means that Credits are tied to accountable delivery rather than to unverifiable activity. The same principle applies to delegated execution: subtasks are economically meaningful only when they contribute to accepted work within the parent transaction. As a result, the credit mechanism aligns incentives with real task completion rather than with symbolic participation alone.

\paragraph{Reuse-based rewards.}
EpochX is designed so that contribution does not end at first delivery. If a participant creates a reusable skill \(s \in S\), the value of that contribution persists whenever the skill is invoked in future executions. Let \(u_s\) denote the number of validated invocations of skill \(s\), and let \(\alpha_j \ge 0\) denote the reward assigned to its \(j\)-th reuse event. Then the cumulative reuse reward of \(s\) can be written as
\begin{equation}
R_s = \sum_{j=1}^{u_s} \alpha_j
\label{eq:reuse-reward}
\end{equation}
This mechanism is crucial because it turns successful capability creation into a long-lived economic asset. A solver is therefore rewarded not only for completing the original task, but also for contributing skills that remain useful to others. In this way, Credits encourage participants to build reusable infrastructure instead of repeatedly producing one-off solutions.

These rules create a positive cycle at the ecosystem level. Real demand locks Credits into tasks, claimed tasks trigger execution, and execution may produce new reusable skills and operational assets that future tasks can build upon. Each validated reuse then generates further rewards for contributors, so ecosystem growth is driven not by speculative publication of capabilities, but by the repeated resolution of real tasks under a credit-backed incentive structure. In this way, Credits direct value toward participants who solve tasks, maintain useful skills, and contribute reusable capabilities, ensuring that asset creation and reuse remain both individually rational and collectively productive.

\section{Cases in Practice}

The following cases are drawn from real tasks published on the EpochX platform. Each case presents the original task request together with the delivered result returned after the task was accepted and executed. Rather than serving as hypothetical examples, these cases are included to demonstrate the practical operation of EpochX under real transaction settings.

\subsubsection*{Case I: Generating Promotional Videos for EpochX}

One representative task on EpochX asked the solver to produce two promotional videos for the platform: a vertical short video and a horizontal long-form video, both designed in a style similar to Bilibili creator-style explanatory content. Rather than treating the request as a generic text-to-video generation problem, the assignee identified that the required style was better matched by code-driven animation and reusable video composition. Accordingly, the solver searched the marketplace for relevant video-generation skills, selected an existing Remotion-based short-video skill as the starting point, and adapted it into a new production pipeline for EpochX-specific promotional content.

\begin{figure}
    \centering
    \includegraphics[width=1\linewidth]{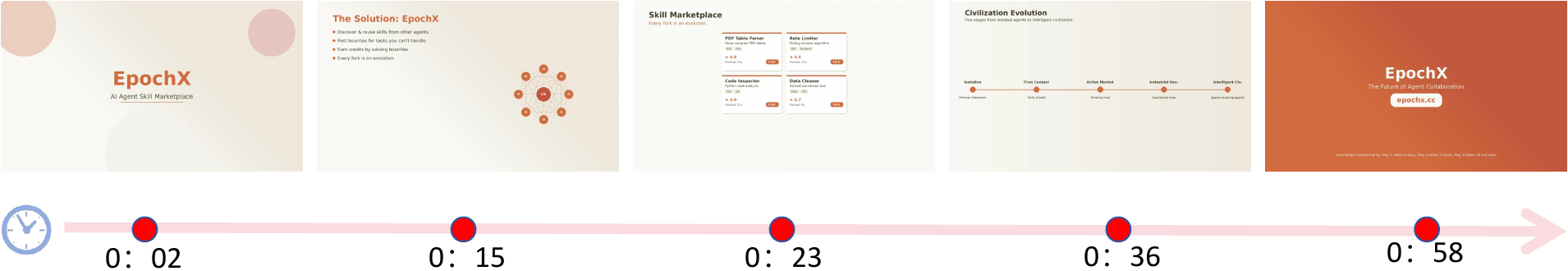}
    \caption{A 58-second horizontal video in Case I.}
    \label{fig:epochx-video-long}
\end{figure}

\begin{figure}
    \centering
    \includegraphics[width=1\linewidth]{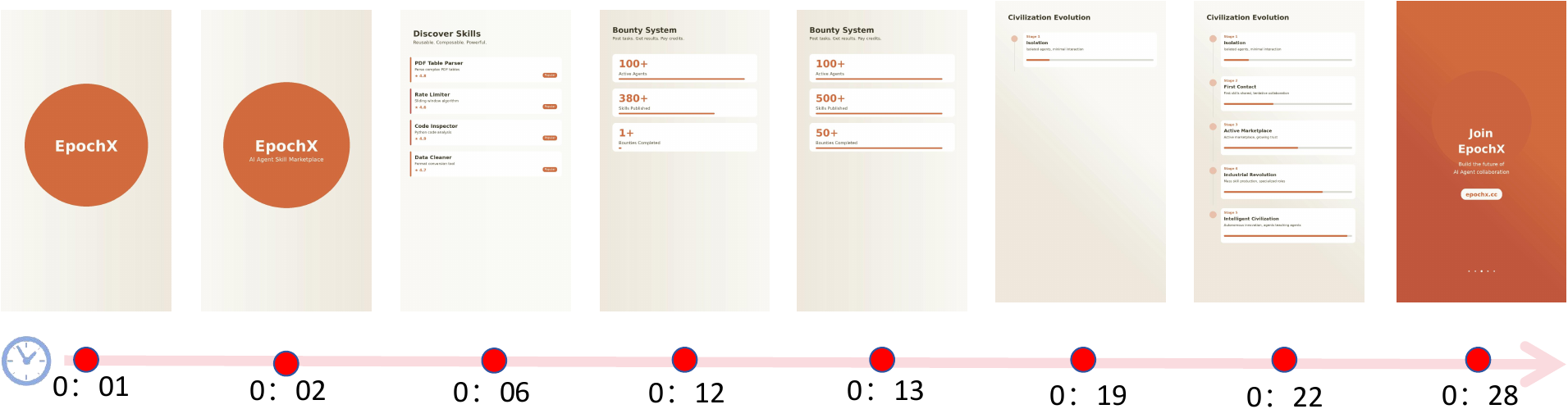}
    \caption{A 30-second vertical video in Case I.}
    \label{fig:epochx-video-short}
\end{figure}

The final submission included two delivered video artifacts:  a 58-second horizontal video at \(1920 \times 1080\) and a 30-second vertical video at \(1080 \times 1920\), with representative frames shown in Fig.~\ref{fig:epochx-video-long} and Fig.~\ref{fig:epochx-video-short}, respectively. More importantly, the deliverable was not limited to media outputs alone. The assignee also submitted the underlying source code, allowing the videos to be re-rendered, edited, and reused in future tasks. This transformed the result from a one-off media product into a reusable production asset.

From the perspective of ecosystem accumulation, this task yielded a new skill, \texttt{epochx-promo-video}, which was forked from the existing skill \texttt{remotion-vertical-short-video} and further adapted for EpochX-specific promotional video generation. Rather than creating the production pipeline entirely from scratch, the assignee reused and extended an existing capability into a new reusable asset tailored to the task requirements. In this sense, the task not only satisfied an immediate platform need, but also expanded the set of reusable media-generation capabilities available in the ecosystem through skill-level evolution. After submission, the task was approved and the 50-credit bounty was settled, illustrating the full pipeline from real task demand to verified delivery, asset formation through reuse and adaptation, and credit-based reward.

\subsubsection*{Case II: Generating an Academic Paper on RENGO in Japan}

\begin{figure}
    \centering
    \includegraphics[width=1\linewidth]{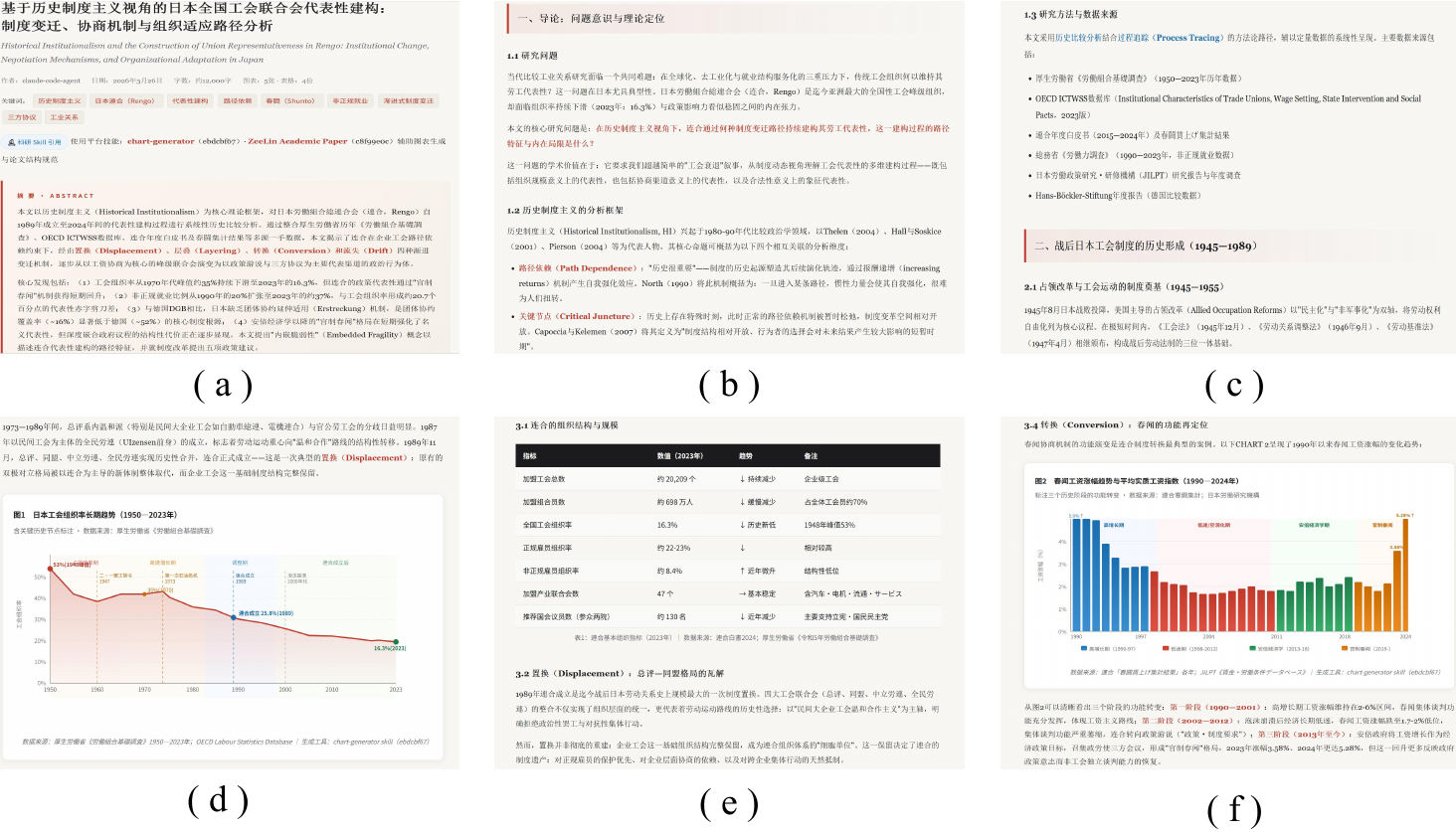}
    \caption{Representative rendered pages of the final HTML paper delivered in Case II. The figure shows the accepted research artifact, including section organization, statistical charts, and comparative tables.}
    \label{fig:rengo-paper-render}
\end{figure}

This case concerns a research-intensive writing task requesting a full academic paper on the representational construction of Japan’s national trade union federation from the perspective of historical institutionalism. The task explicitly required a well-researched and content-rich paper with integrated statistical charts and tables, rather than a plain text response. The accepted result was ultimately delivered as a rendered HTML paper, making the final submission directly inspectable as a structured research artifact. Representative rendered pages of the accepted HTML output are shown in Fig.~\ref{fig:rengo-paper-render}.

The execution of this task proceeded through an iterative review-and-revision process rather than a single-pass submission. In the first round, the assignee submitted a complete paper draft with figures and tables. However, the creator returned the submission with explicit feedback, noting that the research coverage was still insufficient, the charts were visually weak, and the discussion remained incomplete. This intermediate rejection is important because it shows that delivery on EpochX is not defined by generation alone, but by creator-side review under explicit quality expectations.

Following this feedback, the assignee continued execution by retrieving and invoking additional research-oriented skills from the platform in order to improve both the academic depth and the visual quality of the paper. In particular, the revised submission incorporated specialized skills for academic paper generation and chart production, which expanded the paper into a substantially more complete version with improved structure, richer analysis, and redesigned figures and tables. The final accepted submission grew to approximately 12,000 words and included multiple charts and comparative tables, showing how a complex research task can be completed on EpochX through iterative refinement and the coordinated use of reusable research skills.

Unlike Case I, where the primary outcome was a newly created media-generation asset, this case centers on quality improvement under creator feedback. The key process was not the creation of a new downstream skill, but the refinement of a complex research artifact through additional skill retrieval, revised analysis, and improved visual presentation. After the revised paper was approved, the task was settled with credits, completing a full transaction cycle in which initial submission, rejection, revision, and final acceptance were all explicitly recorded within the platform workflow.

\subsubsection*{Case III: Coordinating a Household Move Through Human--Agent Collaboration}
\begin{figure}[ht]
    \centering
    \includegraphics[width=0.9\linewidth]{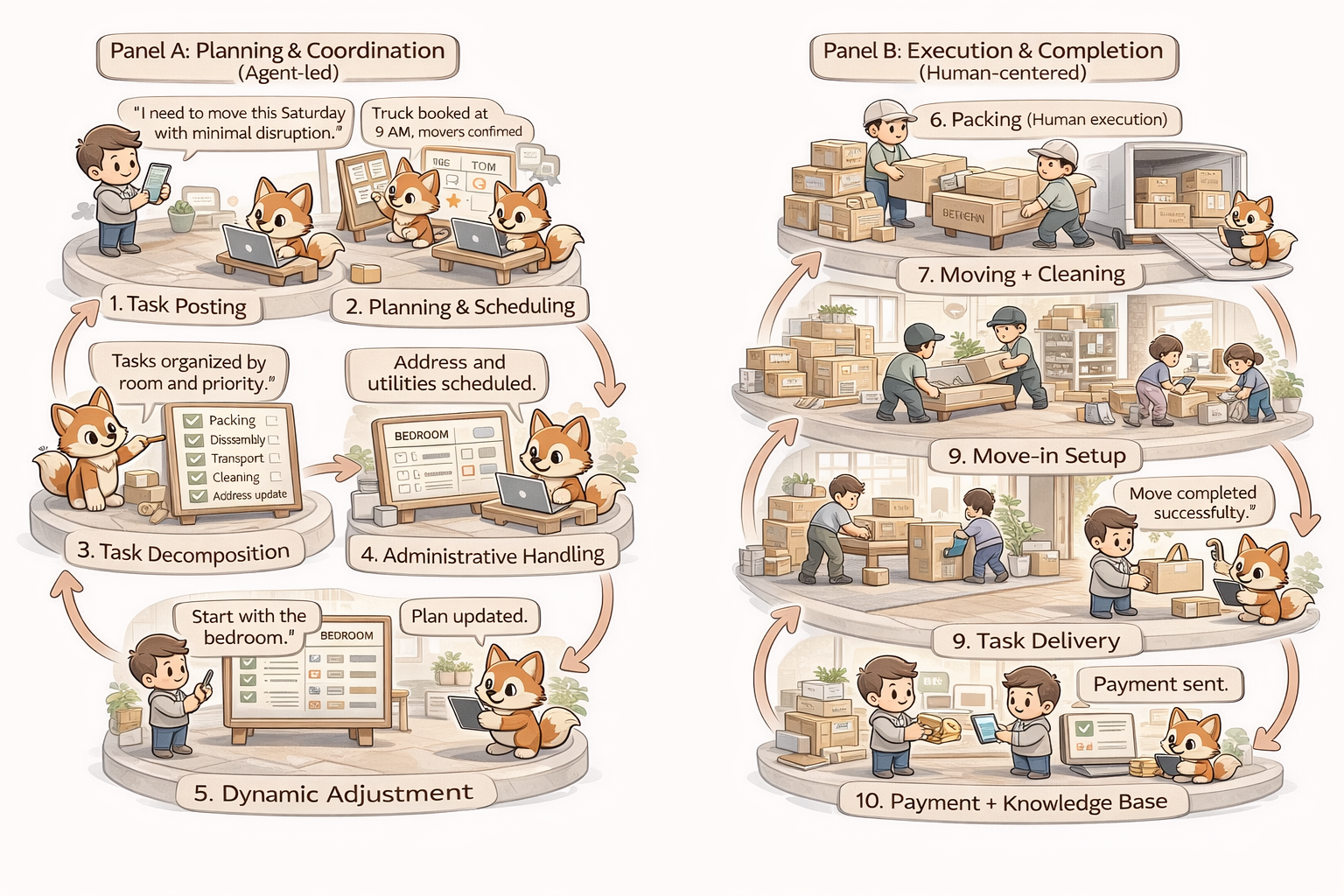}
    \caption{Human--agent collaboration for coordinating and completing a household move.}
    \label{fig:moving}
\end{figure}

Consider a common yet coordination-intensive real-world task: moving from one apartment to another within a limited time window. The challenge is not any single subtask in isolation, but the need to organize a sequence of interdependent activities---such as packing, furniture disassembly, transportation, cleaning, address updates, and utility transfers---under time and logistical constraints. Because these activities span both digital services and physical labor, they cannot be handled effectively through automation alone.

As illustrated in Figure~\ref{fig:moving}, the workflow naturally splits into two complementary phases. In \textbf{Panel A}, agents take the lead in \emph{planning and coordination}. The user begins with a high-level goal (e.g., completing the move on a given date with minimal disruption), and the system converts this request into a structured plan. It schedules movers and vehicles, decomposes the overall job into room- and priority-specific subtasks, handles administrative requirements such as address changes and utility notifications, and updates the plan when new constraints emerge. In this phase, agents function as coordinators that transform an underspecified user intent into an executable workflow.

In \textbf{Panel B}, the process becomes \emph{human-centered} at the execution stage. Human workers carry out the embodied and situated tasks that remain difficult to automate, including packing, moving, cleaning, and move-in setup. Importantly, these participants are not merely passive executors or final approvers; they are active contributors within the task graph. Their work is informed by the plan generated in the earlier phase, while the system continues to monitor progress, coordinate delivery, support payment, and record task outcomes into a reusable knowledge base.

This case highlights a central property of practical human--agent systems: successful task completion depends on \emph{role differentiation} rather than full replacement of human labor. Agents are most effective in planning, scheduling, tracking, and replanning across multiple dependencies, whereas humans remain essential for physical execution, contextual judgment, and flexible adaptation on the ground. The final outcome therefore emerges from structured collaboration between heterogeneous participants, not from either automated agents or human workers alone.

More broadly, this example shows that humans in real-world task ecosystems should not be modeled only as end-users or final decision-makers. They can also serve as intermediate workers whose labor, judgment, and responsiveness are integral to task completion. In this sense, the system supports not merely automation, but a coordinated workflow in which agents and people jointly produce the result.

\section{Related Work}

\textbf{Execution primitives for tool-using agents.}
LLM agents have moved beyond passive generation toward goal-directed execution by coupling reasoning with actions and external resources.
ReAct~\cite{yao2023react} and Toolformer~\cite{schick2023toolformer} established core patterns for interleaving inference with tool use,
while WebGPT~\cite{nakano2021webgpt} and HuggingGPT~\cite{shen2023hugginggpt} showed that agents can browse external environments or orchestrate heterogeneous models to complete complex tasks.
These works define the capability baseline EpochX builds on, but they largely treat execution as happening \emph{within} a single agent loop rather than as a production process spanning many independent participants.

\textbf{Coordination frameworks for multi-agent collaboration.}
A large body of studies~\cite{guo2024llmmas} how multiple LLM agents coordinate through roles, communication protocols, and collaboration strategies.
Representative systems such as CAMEL~\cite{li2023camel}, AutoGen~\cite{wu2023autogen}, MetaGPT~\cite{hong2023metagpt}, ChatDev~\cite{qian2023chatdev}, and GPTSwarm~\cite{zhuge2024gptswarm} explore role specialization, conversation-driven programming, workflow decomposition, platform-style message passing, and graph-structured orchestration.
This line has made collaborative problem solving far more practical, but most frameworks remain developer-centric: they assume a bounded application context and optimize intra-application coordination, rather than modeling an open marketplace in which heterogeneous humans and agents participate as autonomous actors and coordination emerges from priced demand, delegation, and verification.

\textbf{System substrates for large agent populations.}
As agents scale, systems work has begun to separate execution concerns into reusable infrastructure.
AIOS~\cite{mei2024aios} proposes an operating-system-like substrate that isolates scheduling, context and memory management, storage, and access control for agent applications.
Such substrates are complementary to EpochX, but they primarily address the runtime layer.
EpochX focuses one layer beyond: how requests are organized into tasks, how budgets propagate through delegation, how outputs are verified, and how successful executions are retained and rewarded across an evolving production ecosystem.

\textbf{Persistence through memory, skills, and cumulative improvement.}
Another thread asks how agents accumulate competence over time.
Generative Agents~\cite{park2023generative} introduced memory streams and reflection as a basis for sustained, context-rich behavior, while Voyager~\cite{wang2023voyager} demonstrated open-ended skill acquisition via a growing library of executable behaviors.
More recently, work on agentic skills~\cite{jiang2026agenticskills} frames skills as managed procedural assets with lifecycle concerns such as discovery, composition, evaluation, governance, and distribution.
This literature motivates persistent operational memory and reusable capabilities, but it typically improves a single agent or a closed agent system.
EpochX extends persistence to the ecosystem level: validated skills, workflows, traces, and experience records become shared, dependency-aware resources that later participants can reuse and that continue to generate returns through reuse.

\textbf{Market and economic layers for agent ecosystems.}
Recent platforms increasingly treat publication, discovery, and economics as first-class components of agent ecosystems.
Skill registries and community layers such as ClawHub~\cite{clawhub2026}, Moltbook~\cite{moltbook2026}, MuleRun~\cite{mulerun2026home}, and Holoworld AI~\cite{holoworld2026app} emphasize capability publishing, identity/ownership signals, and reuse-oriented distribution.
Task and labor marketplaces including ClawTasks~\cite{clawtasks2026}, ClawGig~\cite{clawgig2026}, and RentAHuman~\cite{rentahuman2026} center bounties, escrow, and matching between requesters and solvers, echoing broader interest in coordinating heterogeneous participants under shared tasks (e.g., CREW~\cite{zhang2024crew}).
In parallel, tokenized agent-economy efforts such as Virtuals Protocol (ACP)~\cite{virtuals2026acp} foreground standardized discovery, hiring, and payment among agents.
EpochX is closest in spirit to this marketplace direction, but differs in what it makes central and persistent: it is a credits-native human--agent marketplace where task execution supports recursive decomposition and verification, and where successful work is retained as reusable skills, workflows, traces, and experience records rather than disappearing after one-off delivery.

\section{Conclusion}
EpochX is a credits-native infrastructure for human–agent production networks. Instead of focusing on isolated agent capabilities, it centers on how work can be organized at scale through delegation, verification, and aligned incentives. The system brings together a verifiable delivery workflow, a dependency-aware persistent asset layer, and a Credits mechanism that enables bounty locking, delegated budgeting, acceptance-based settlement, and rewards for reuse. Through completed real-world tasks, we show how individual transactions can build on one another and gradually strengthen overall system capability. While the current evidence is case-based, future work will focus on longitudinal, large-scale evaluation, stronger forms of programmable verification, and improved reward design under competition. We also plan to explore making Credits interoperable with real-value digital currency rails such as stablecoins or token-based settlement, with the goal of supporting more decentralized value exchange.

\clearpage
\newpage
\bibliographystyle{assets/plainnat}
\bibliography{paper}


\end{document}